# Parallel, Self Organizing, Consensus Neural Networks

Homayoun Valafar
University of Georgia, CCRC
Athens, GA 30602
Ph: (706) 542-4401
Email: homayoun@ccrc.uga.edu

Faramarz Valafar
University of Georgia, CCRC
Athens, GA 30602
Ph: (706) 542-4436
Email: faramarz@ccrc.uga.edu

Okan Ersoy
Purdue University, MSEE
West Lafayette, IN 47907
Ph: (317) 494-6162
Email: ersoy@ecn.purdue.edu

**Abstract**

A new neural network architecture (PSCNN) is developed to improve performance and speed of such networks. The architecture has all the advantages of the previous models such as self-organization and possesses some other superior characteristics such as input parallelism and decision making based on consensus. Due to the properties of this network, it was studied with respect to implementation on a Parallel Processor (Ncube Machine) as well as a regular sequential machine. The architecture self organizes its own modules in a way to maximize performance. Since it is completely parallel, both recall and learning procedures are very fast. The performance of the network was compared to the Back propagation networks in problems of language perception, remote sensing and binary logic (Exclusive-Or). PSCNN showed superior performance in all cases studied.

## Introduction

*Parallel, Self-organizing, Consensual Neural Network* (*PSCNN*) is an alternative to the conventional cascaded neural networks. This network is a predecessor of *Hierarchical Neural Networks* [1] . *PSCNN* offers a better performance[2], a faster algorithm that can even be considered for real time execution, self organization for optimal performance and finally a better and closer emulation of human brain for perception experiments such as speech and vision.

*PSCNN* is an architecture which will create a purely parallel environment for the operation of neurons. This architecture will not only simulate the concept of modularity in the human brain but it will also self organize in the number of stages needed in order to achieve an optimal performance.

*PSCNN* is a unification of several smaller and more primitive modules. Each one of these modules can be chosen at will, but, in this thesis, it is a fully connected, feed-forward, single stage network. These modules are designed in this thesis as completely independent of one another during the training and recall procedures. Therefore, all of the modules can be trained simultaneously and independently. This feature allows a highly parallel operation during the training procedure, which it has not been possible in the past. Further more, this feature of *PSCNN* allows learning of a massive amount of data with a very high performance in relatively a short time. An example of a 4 stage *PSCNN* is shown Figure 1. In this figure NLT stands for Nonlinear Transformation.

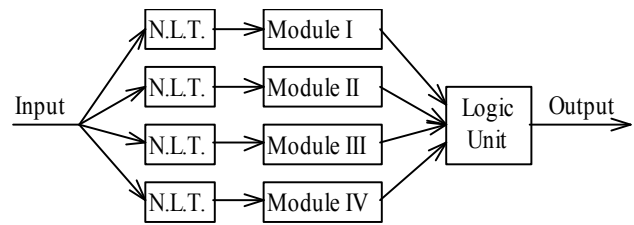

**Figure 1. A diagram of a four module PSCNN.**

## Input and Output to PSCCN Networks

The inputs to each *PSCNN* module is the original training set, transformed non-linearly several times (the number of transformations will depend on the module number). These transformations can be of any kind (as long as they are one-to-one, on-to and nonlinear transformations) such as FFT, DFT followed by point-wise non-linearities, or even a simple binary operation such as perfect shuffling, two's or one's complement [3]. In this research, a more drastic binary operation was used, namely Gray code of a binary code. Although the input to each module is radically changed, the desired output of each module remains the same as the original desired output. All the initial weights will be selected as random numbers.

Gray code [4] is a kind of binary representation of numbers like Two's complement or others. The advantage of gray coding is that numbers which appear to be very similar to each other in the One's or Two's complement representation will be very dissimilar in gray code. This will allow the easy recognition of similar and hard to detect vectors. Gray code transformation has an additional property of rotation of only a part of the binary space. This nonlinear rotation of space is very effective in reducing the distances between far points while increasing the distances of near points in the original binary space. This advantage of Gray code transformation is fully explored and illustrated in XOR problem.

## Training of PSCNN Networks

The core learning algorithm of any neural network may consist of one of several available methods. Delta rule (conjugate descent) was selected as the core training algorithm in our experiments. Eventhough there exist more powerful and effective minimization (learning) algorithms we chose delta rule for the following two reasons: first to more effectively establish the success of this algorithm/architecture, and secondly to be able to compare the results to other results obtained as the result of delta rule learning. Therefore, one clear method of improving the performance of such networks is the employment of a more powerful minimization algorithm.

Training of each module starts with a regular update of weights based on simple delta rule learning algorithm. The training is performed on the locally transformed data for each module. The training of each module is often limited to a certain number of epochs and not allowed to converge to a local or global minimum point. This is again to establish the effectiveness of this network. After the terminated training, the architecture selects the required number of modules and discards the remaining if any in order to achieve the required performance. Due to the practical constraints, to prevent the size of the network to exceed a certain limit, an upper limit to the number of modules that *PSCNN* can create is established ahead of time. This upper limit should reflect the computational resources that are available to this algorithm. Once the required modules have been selected, then the *accept* or *reject* boundaries will be determined (explained in next section). At this point the training of the system is complete and thus testing or recall may start.

## Adjustment of Output Boundaries

Output boundaries are defined for each output neuron in order to allow each output of each module to express its level of certainty in regards to the classification task. If a neural network is not given the option of abstaining in the task of decision making, then it is likely that it will produce the wrong output. As a result, the correct classification produced by the minority of modules is overlooked by the incorrect classification produced by the majority. Therefore, by allowing the option of making no decision or an unsure decision, the likelihood of the emergence of the rarely correct and confident classification results will be increased.

Each output neuron in this network (all modules) is designed to establish 5 different output regions and produce and output corresponding to that region. Each region will carry a different weight in establishing the final, combined decision of the entire network. The 5 output regions are as following and are illustrated in Figure 2.

- Definite zero (output = -1).
- Indefinite zero (output = -0.5).
- No decision (output = 0).
- Indefinite one (output = 0.5).
- Definite one (output = 1).

The thresholds for each of these regions is established at the end of the training. After the termination of training, these thresholds can be established based on different rules. The following describes the methods of establishing these thresholds in the experiments that were conducted in this research.

- The maximum number below which no false 0 outputs are produced is the definite 0 threshold.
- The maximum number above which no correct 0 outputs are produced is the upper threshold for indefinite 0 region.
- The minimum number above which no false 1 outputs are produced is the definite 1 threshold.
- The minimum number below which no correct 1 outputs are produces is the lower threshold for indefinite 1 regions.
- Any remaining region between the indefinite regions is the undecided region.

Note that by definition, it is not possible for the two definite regions to overlap, however it is possible to have overlapping indefinite regions. In this scenario, the common region is marked to be the undecided region.

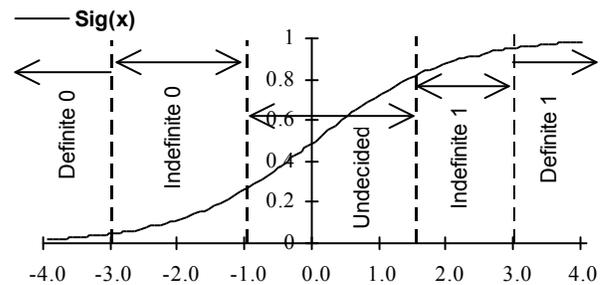

**Figure 2. An example of 5 output regions for any given output neuron.**

## Output of the Network

The logic unit determines the final output of the network by averaging the outputs of all of the modules. The highest output is considered to be the final output. Other methods of determining the final classification result based on the results of the modules can be implemented. For example to further eliminate possible ambiguity in the decision one can modify the output of each neuron to be determined by the following equation.

$$O = O_r + \frac{D_r}{\|r\|}$$

Where O is the final output, $O_r$ is the output corresponding to region r. $\Delta_r$ is the distance that the activation level falls into the region and $\|r\|$ is the total size of the region. This modified output determination allows to resolution of competition among the output neurons of different classes. For example if two neurons have the output of one but one is 0.4 units into the definite and the other one is 0.35, then the first neuron will be a winner over the second one.

## Testing of PSCNN Networks

During the testing, the test vector is transformed with the proper transformation specific to each module and then fed into the module. Each module classifies the input. A final decision-making module, which can be a small computer, analog circuit or even a logic circuit, gathers the responses from all the modules and makes a final decision based on majority voting.

## Experimental Results

### X-Or Problem

Several sets of data were tested on this network to study its performance in comparison to backpropagation networks. A simple but yet difficult problem, namely the exclusive-or problem (XOR) was tested first to discover the behavior and operation of this network. This problem has been studied thoroughly by scientist and engineers during the past several years using various networks.

Study of the solution of PSCNN network to the Exclusive-Or problem helps to understand the functional mechanism of this algorithm. The first step in the study of the PSCNN's solution is the examination of the original problem space illustrated in Figure 3. It is a common knowledge that a single stage neural network is not capable of solving this problem. The solution to this problem requires at least a two stage network and with backpropagation selected as the learning algorithm, this problem can be solved with step size of less than 0.75, 2-40 hidden neurons and 558-6857 epochs of training [Error: Reference source not found]. In contrast, PSCNN succeeded in solving this problem in as few as two modules in which the step sizes were between 0.2 and 0.9 with the training epochs less than 50. PSCNN's solution to this problem is illustrated in Figure 4a&b.

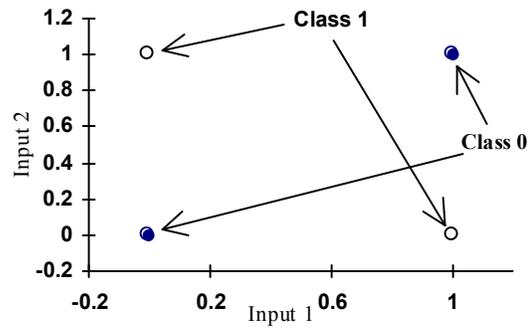

Figure 3. X-Or problem space.

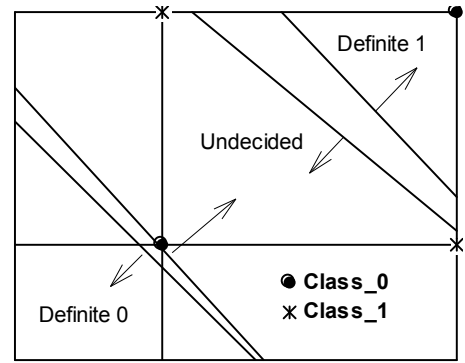

Figure 4a. The first module solution of PSCNN to the XOR problem.

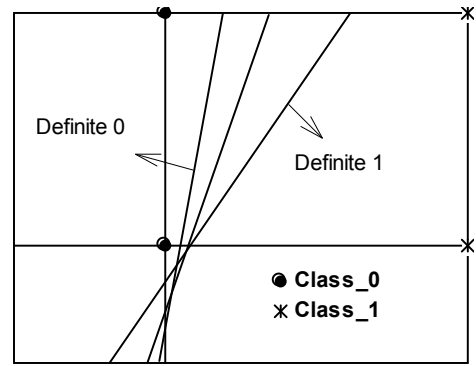

Figure 4b. The second module solution of PSCNN to the XOR problem.

### Remote Sensing Problem:

The data set for this experiment is based on Multispectral Earth observational remote sensing data called Flight Line C1 (FLC1). The geographic location of the FLC1 is the southern part of the Tippecanoe county, Indiana. This multispectral image was collected with an airborne scanner in June 1966 at noon time. The FLC1 consists of 12 band signals with each band corresponding to a different farm product. Only 4 spectral bands (farm products) out of the

12 are used in this experiment. These 4 farm products consisted of alfalfa, corn, oats, and red clover.

The purpose of this experiment is to identify the farm product based on the observed shade of gray. The length of the input vector for each PSCNN module is 64 and the length of the output vector is 4. The learning rate ρ is set to 0.9/k where k is the number of iterations. The initial weights were randomly generated as numbers between -2.5 and 2.5. The results of these experiments are shown in Figure 5. This figure contains the results of the performance of PSCNN after 10, 50 and 500 iterations versus the number of modules allocated to the problem. These are the results of 10, 50 and 500 iterations during the learning respectively. Note that the best performance obtained from a two stage neural network (with optimal number of hidden neurons) trained with back-propagation was 88%.

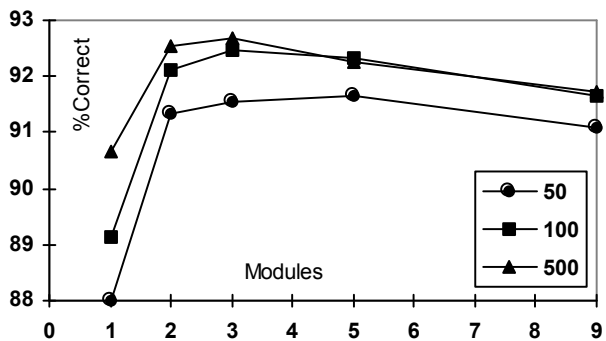

**Figure 5. Performance of PSCNN for the remote sensing problem as a function of number of modules.**

## Conclusion and Discussion

PSCNN offers many attractive features for engineers as well as scientists in other disciplines. The following is a list of some the advantages that PSCNN offers:

- No need for guessing the number of hidden neurons since it is possible to achieve the same performance as s multi-stage network with several single stage networks.

- Much more forgiving in selecting the step size since the discovery of a near minimum optimal point is sufficient for each module.

- It is possible to extract more detailed information regarding the spatial geometry of the clusters in space by examining the success of each module and its transformation in the task of classification.

- PSCNN can be implemented on a parallel machine to utilize its parallel functionality for speedup of classification.

The results of XOR problem demonstrates that how powerful and effective this network is in combining partial solutions in order to provide a global solution to the problem. Another feature of PSCNN is its ability to self-organize. This will allow a network to take the optimal path towards the maximum performance.